\documentclass{article} 
\usepackage{iclr2024_conference,times}

\usepackage{amsmath,amsfonts,bm}

\def\eqref#1{equation~\ref{#1}}

\def\1{\bm{1}}

\DeclareMathAlphabet{\mathsfit}{\encodingdefault}{\sfdefault}{m}{sl}
\SetMathAlphabet{\mathsfit}{bold}{\encodingdefault}{\sfdefault}{bx}{n}

\usepackage{graphicx}
\usepackage{hyperref}
\usepackage{longtable}
\usepackage{multirow}
\usepackage{url}

\newcommand\myparagraph[1]{\vspace{6pt}\noindent\textbf{#1}\quad}

\title{Squeezing lemons with hammers: An evaluation of AutoML and tabular deep learning for data-scarce classification applications}

\author{Ricardo Knauer and Erik Rodner\\
KI-Werkstatt, University of Applied Sciences Berlin\\
\url{https://kiwerkstatt.htw-berlin.de}\\
\texttt{ricardo.knauer@htw-berlin.de}
}

\iclrfinalcopy
\begin{document}

\maketitle

\begin{abstract}
Many industry verticals are confronted with small-sized tabular data. In this low-data regime, it is currently unclear whether the best performance can be expected from simple baselines, or more complex machine learning approaches that leverage meta-learning and ensembling. On 44 tabular classification datasets with sample sizes $\leq$ 500, we find that L2-regularized logistic regression performs similar to state-of-the-art automated machine learning (AutoML) frameworks (AutoPrognosis, AutoGluon) and off-the-shelf deep neural networks (TabPFN, HyperFast) on the majority of the benchmark datasets. We therefore recommend to consider logistic regression as the first choice for data-scarce applications with tabular data and provide practitioners with best practices for further method selection.
\end{abstract}

\section{Introduction}

Machine learning algorithms thrive on data \citep{banko2001scaling,halevy2009unreasonable}. In particular, the availability of massive text corpora crawled from the web has been the foundation for large language models such as GPT-3, and has given rise to astounding emergent abilities such as in-context learning \citep{brown2020language}. In other domains, though, data are not as easy to come by. In the clinical diagnostic setting, for instance, medical conditions can be too rare to collect many instances; in the clinical prognostic setting, patients can be lost at future follow-ups. Under these circumstances, much simpler predictive models have traditionally been advocated \citep{moons2015transparent,steyerberg2019clinical}. This preference for simpler models is supported by a systematic review that indeed did not find a difference in discriminative performance between logistic regression and more complex machine learning models on average \citep{christodoulou2019systematic}. But also beyond the clinical domain, it has been recommended to try simpler baselines first before switching to more complex models \citep{mcelfresh2023neural}. In the largest tabular data analysis to date, the overall best performance was achieved by a meta-trained transformer ensemble \citep{hollmann2023tabpfn}; perhaps more importantly, simpler models like logistic regression performed on par with or better than more complex models on 31\% of the datasets \citep{mcelfresh2023neural}. However, with median sample sizes $>$ 1000, these large-scale analyses only provide insufficient guidance for practitioners to select the most suitable machine learning approach when data is limited. In particular, meta-learning and ensembling appear to be well-suited for the low-data regime \citep{hollmann2023tabpfn,mcelfresh2023neural}, but it remains unclear if or when approaches that leverage these strategies, such as automated machine learning (AutoML) \citep{alaa2018autoprognosis,erickson2020autogluon,imrie2023autoprognosis,salinas2023tabrepo} or pretrained deep neural networks \citep{bonet2023hyperfast,hollmann2023tabpfn,muller2023mothernet}, indeed offer an appreciable performance benefit for data-scarce applications (e.g., with sample sizes $\leq$ 500).

Our contributions are as follows:
\begin{enumerate}
\item We perform the \textbf{first extensive comparison of state-of-the-art machine learning methods, i.e., AutoML frameworks and off-the-shelf deep learning models, against a simple L2-regularized logistic regression baseline on 44 (very-)small-sized tabular datasets}.

\newpage In summary, we find that L2-regularized logistic regression shows a similar discriminative performance to more complex models on 55\% of the datasets (Sect.~\ref{sec:results}), and that correlations of simple dataset meta-features with performance differences are only small (Sect.~\ref{sec:metafeature}). We therefore recommend to use more complex models only if the performance with logistic regression is considered insufficient for the application at hand, in line with \citet{mcelfresh2023neural}.

\item We provide practitioners with the \textbf{best L2-regularization hyperparameter obtained on each dataset that can be used for meta-learning in data-scarce applications} (Sect.~\ref{sec:results}).
\end{enumerate}

\section{Related work} \label{sec:relatedwork}

\myparagraph{Challenges in the low-data regime} Data-scarce applications pose unique challenges for machine learning, especially when combined with a relatively low number of labels in the smaller class per feature, or outcome events per variable (EPV). Clearly, overfitting is a concern for more complex models. Perhaps less obviously, finding good hyperparameter settings also becomes more difficult when data is limited. Low sample sizes produce cross-validation folds that are even smaller than the original sample, which may in turn be too small to adequately represent both the original sample and the population of interest, causing suboptimal hyperparameters. In case of regularized logistic regression, for instance, it has been shown that penalty strengths are generally estimated with increasing uncertainty as the sample size or EPV decreases, which can increase the between-sample variability of the predictive performance \citep{riley2021penalization,vsinkovec2021tune,van2020regression}. This implies that data-driven hyperparameter optimization methods may fail to increase, or may even decrease, the predictive performance for an individual dataset.

\myparagraph{Machine learning solutions} The traditional approach to attenuate overfitting and facilitate the hyperparameter search process in the low-data regime is to use regularized logistic regression with a constrained search space \citep{moons2015transparent,steyerberg2019clinical}; recent solutions like AutoPrognosis, AutoGluon, TabPFN, and HyperFast leverage more complex models in combination with meta-learning and ensembling. AutoPrognosis \citep{alaa2018autoprognosis,imrie2023autoprognosis} is an AutoML framework that automates building end-to-end machine learning pipelines, including preprocessing, model, and hyperparameter selection. Both logistic regression and tree ensembles are considered for algorithm search. Its default pipeline optimization procedure is initialized using meta-learned hyperparameters from external datasets, and a pipeline ensemble is constructed following the pipeline search process. In contrast to AutoPrognosis, AutoGluon \citep{erickson2020autogluon,salinas2023tabrepo} is an AutoML framework that meta-learns a model hyperparameter portfolio from external datasets and different random seeds (zero-shot hyperparameter optimization), and considers k-nearest neighbors, tree ensembles, and neural networks for selection. It focuses the training time budget on ensembling instead of further hyperparameter optimization. Recent off-the-shelf deep learning models, sometimes referred to as tabular foundation models \citep{muller2023mothernet}, also prioritize ensembling over hyperparameter tuning. TabPFN is a meta-trained transformer ensemble that performs in-context learning on tabular data \citep{hollmann2023tabpfn}. HyperFast meta-trains a hypernetwork with external datasets, uses the pretrained hypernetwork to generate smaller, task-specific main networks with the actual training dataset, and optionally fine-tunes and ensembles the main networks \citep{bonet2023hyperfast,muller2023mothernet}. Overall, AutoGluon, TabPFN, and HyperFast have generally outperformed (L1-, L2- or unregularized) logistic regression on 40 small- to medium-sized tabular datasets (median sample size 1241, range [500, 45,211]) \citep{bonet2023hyperfast,hollmann2023tabpfn,muller2023mothernet,puri2023semi}. In the next section, we extend the prior results to the data-scarce setting, by evaluating AutoPrognosis, AutoGluon, TabPFN, as well as HyperFast against a simple L2-regularized logistic regression baseline on 44 smaller-sized tabular datasets (median sample size 204, range [32, 500]).

\section{Experiments}

In the following, we describe our experimental setup to assess two AutoML frameworks and two pretrained deep neural networks against L2-regularized logistic regression (Sect.~\ref{sec:relatedwork}) on 44 (very-) small-sized tabular datasets with a 1h runtime limit on 8 vCPU cores with 32GiB memory. We then present our experimental results together with the best L2-regularization hyperparameter setting for each dataset. Additionally, we conduct a meta-feature analysis to evaluate under which conditions, i.e., dataset properties, AutoML and deep learning methods perform better than logistic regression. Overall, we observe a similar discriminative performance for L2-regularized logistic regression to more complex models on 55\% of the datasets. Furthermore, correlations of simple dataset meta-features with performance differences are only small. We therefore recommend to generally start predictive modeling in the low-data regime with a logistic regression baseline first, and to use more complex approaches only if the obtained performance is insufficient, echoing the conclusions of \citet{mcelfresh2023neural}.

\subsection{Experimental setup}

\myparagraph{Datasets and methods} We used all binary classification datasets with sample sizes $\leq$ 500 from Penn Machine Learning Benchmarks (PMLB) \citep{olson2017pmlb,romano2022pmlb}, yielding 44 curated tabular datasets in total (median sample size 204, range [32, 500]). We shuffled all datasets, encoded all labels to \{0, 1\} for AutoML and deep learning methods and to \{-1, 1\} for L2-regularized logistic regression, subsampled the clean1 dataset to 100 features for TabPFN \citep{feuer2023scaling}, and "min-max" scaled all features for logistic regression. We used AutoPrognosis 0.1.21 with the default settings, AutoGluon 1.0.0 with the "best quality" preset, and TabPFN 0.1.9 with 32 ensemble members \citep{hollmann2023tabpfn}. For HyperFast 0.1.3, we observed runtimes exceeding our 1h budget using the default settings, so we decided not to fine-tune the main networks, but instead to increase the number of ensemble members to 32 \citep{bonet2023hyperfast,hollmann2023tabpfn}. For L2-regularized logistic regression, we used a continuous conic formulation \citep{knauer2023cost} with the default optimality gaps via MOSEK 10 \citep{aps2023mosek} and JuMP 1.4.0 \citep{dunning2017jump} from Julia 1.8.3. 

\myparagraph{Evaluation metrics} For each dataset and method, we measured the training area under the receiver operating characteristic curve (AUC) on the whole data with a 1h runtime limit, as well as the mean test AUC via a stratified, 3-fold cross-validation procedure with a 1h runtime limit per fold (i.e., 4h per dataset in total). For L2-regularized logistic regression, we instead used cross-validation to tune the L2-regularization hyperparameter $\lambda$. We set $\lambda$ to one value of the geometric sequence [0.5, 0.1, 0.02, 0.004], used the deviance to find the best hyperparameter value $\lambda^*$, and tracked the training AUC with $\lambda^*$. To compute the mean test AUC, we used a nested cross-validation procedure \citep{knauer2023cost}.

\subsection{Experimental results} \label{sec:results}

\begin{figure}[t]
\centering
\includegraphics[width=0.70\textwidth]{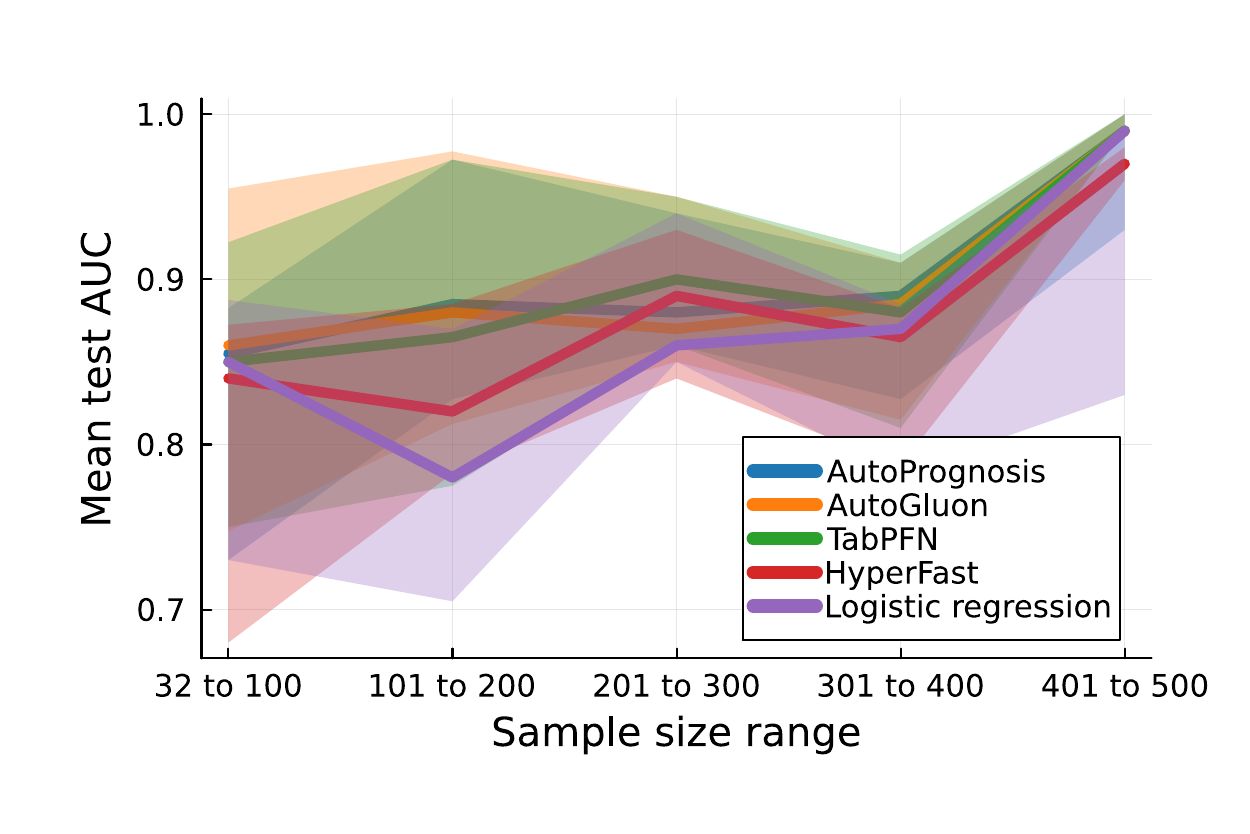}
\caption{Mean test AUCs with their medians and interquartile ranges shown across sample size ranges in steps of 100 for AutoML, deep learning, and logistic regression.}
\label{fig:results_a}
\end{figure}

\begin{figure}[t]
\centering
\includegraphics[width=0.34\textwidth]{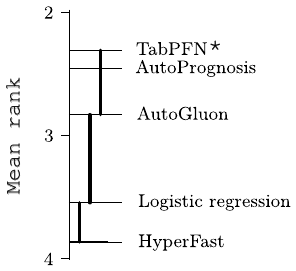}
\caption{Critical difference diagram to detect pairwise mean test AUC differences, based on the Holm-adjusted Wilcoxon signed-rank test \citep{demsar2006statistical,benavoli2016should}, for AutoML, deep learning, and logistic regression. Approaches that are not statistically different at the 0.05 significance level are connected by a bold vertical bar. $^\mathbf{\star}$Note that TabPFN results are biased (Sect.~\ref{sec:results}).}
\label{fig:results_b}
\end{figure}

AutoPrognosis, AutoGluon, TabPFN, HyperFast, and logistic regression perform relatively similar at different sample sizes, with a trend towards better performance when the sample size increases for all methods (Fig.~\ref{fig:results_a}). The largest difference occurs at sample sizes from 101 to 200 where HyperFast and logistic regression only achieve a median discriminative performance of 0.82 and 0.78, respectively. Similar to \citet{mcelfresh2023neural}, though, we find that each approach wins on at least one dataset and looses on at least one other dataset (Table~\ref{table:results} in Appendix~\ref{sec:appendix}). Logistic regression performs on par with or better than the best AutoML or deep learning method on 16\% of the datasets, possibly because the latter regularly overfit, especially with smaller sample sizes. On 34\% of the datasets, logistic regression performs within 1\% of the best AutoML or deep learning approach; on 48\% of the datasets within 2\% of the best approach; and on 55\% of the datasets within 3\%. Interestingly, even AutoML methods that consider logistic regression for algorithm selection (i.e., AutoPrognosis) may be outperformed by logistic regression alone (e.g., on the backache dataset), again possibly due to overfitting. From a statistical perspective, Fig.~\ref{fig:results_b} shows that AutoPrognosis and TabPFN achieve a better rank than logistic regression (p $<$ 0.05), whereas AutoGluon and HyperFast are not different from a simple logistic regression baseline (p $>$ 0.05). It must be noted, though, that TabPFN's prior was developed on 45\% of the evaluated datasets \citep{hollmann2023tabpfn}. The performance estimates for TabPFN are therefore likely to be overoptimistic. Finally, the best L2-regularization hyperparameters are shown in Table~\ref{table:results} in Appendix~\ref{sec:appendix}, and they can be readily used for meta-learning, e.g., by leveraging the most similar PMLB dataset(s) for zero- or few-shot hyperparameter optimization \citep{feurer2015initializing,reif2012meta}.

\subsection{Meta-feature analysis} \label{sec:metafeature}

In this section, we extract meta-features from our datasets and compute associations of these meta-features with performance differences between each AutoML / deep learning method and logistic regression. This allows us to find out when more complex machine learning models succeed and derive best practices for method selection.

\myparagraph{Meta-feature extraction} We extracted meta-features for each dataset using PyMFE 0.4.3 for all available meta-feature groups \citep{alcobacca2020mfe}, including but not limited to simple meta-features (such as the sample size and EPV (Sect.~\ref{sec:relatedwork})), statistical meta-features (such as the feature means and standard deviations), clustering meta-features (such as the mean silhouette value and Dunn index), information-theoretic meta-features (such as the feature and class entropies), model-based meta-features (such as the feature importances and number of leaf nodes in a decision tree), as well as complexity and landmarking meta-features (such as the predictive performance of a linear support vector machine and k-nearest neighbors classifier). Some meta-features, such as the feature importance measure, contained multiple values. Therefore, we aggregated these values with all available summary functions in PyMFE, including but not limited to the mean, standard deviation, and frequency in a particular histogram bin, yielding 3932 meta-features in total.

\begin{figure}[t]
\centering
\includegraphics[width=0.87\textwidth]{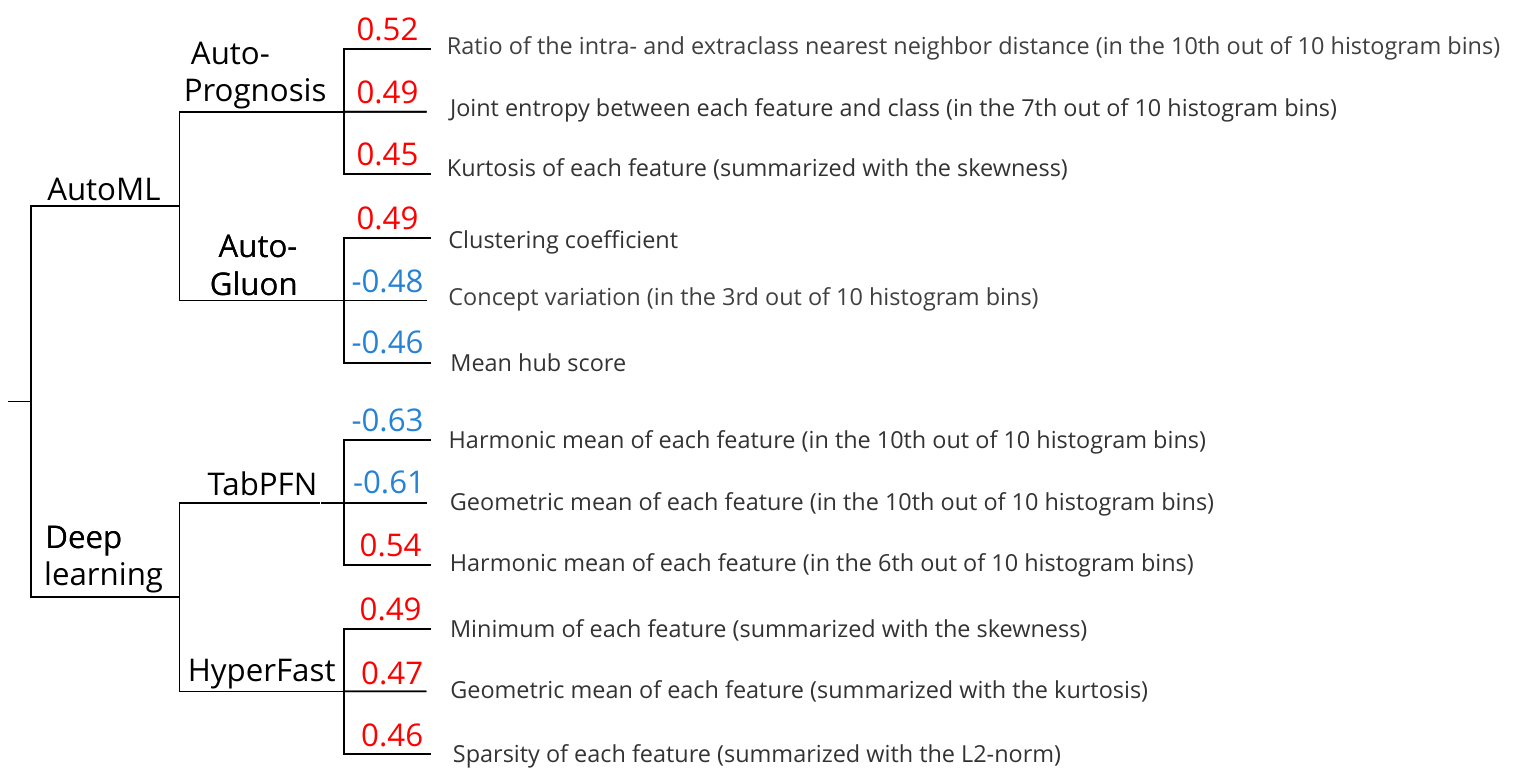}
\caption{Top-3 meta-features according to their relationship with the mean test AUC differences between each AutoML / deep learning method and logistic regression, based on absolute Spearman rank correlations. Positive correlation coefficients are shown in red, negative coefficients in blue.}
\label{fig:results_c}
\end{figure}

\myparagraph{Correlation analysis} Fig.~\ref{fig:results_c} shows the top-3 correlations of our meta-features with the performance differences between each AutoML / deep learning approach and a logistic regression baseline. The most discriminating meta-feature for AutoPrognosis is the ratio of the intra- and extraclass nearest neighbor distance, for AutoGluon the clustering coefficient. Both complexity measures are larger for harder classification problems \citep{lorena2019complex}, and show a positive relationship with the performance differences. AutoML methods therefore appear to be better suited for more complex datasets that need more complex classifiers. For TabPFN and HyperFast, the most discriminating meta-features are summary scores from the feature distribution, i.e., the harmonic mean and minimum of each feature. Pretrained deep neural networks may therefore be better suited for datasets with feature distributions similar to their meta-training datasets. Of note, simple meta-features show much smaller correlations coefficients than those depicted in Fig.~\ref{fig:results_c}. The correlation coefficients of the sample size with the performance differences are only -0.09, -0.09, -0.02, and 0.08 for AutoPrognosis, AutoGluon, TabPFN, and HyperFast, respectively; the correlation coefficients of the EPV are only -0.30, -0.36, -0.33, and -0.20. Simple meta-features therefore appear not to be a good guide when to choose more complex machine learning approaches over a logistic regression baseline for data-scarce applications.

\section{Conclusion}

The low-data regime presents challenges for machine learning. In particular, overfitting and suboptimal hyperparameter settings can make it more difficult to achieve a good performance when data is limited. In this work, we evaluated a simple L2-regularized logistic regression baseline against state-of-the-art AutoML and deep learning approaches on 44 (very-)small-sized tabular datasets, and found that the discriminative performance is in fact similar 55\% of the time. What is more, simple dataset meta-features like the sample size or EPV are not sufficiently correlated with the performance differences to select the most suitable machine learning approach before actually trying it out (at least when considering a bivariate monotonic correlation measure like we have). For that reason, we recommend not to "squeeze lemons with hammers", but to use logistic regression for data-scarce applications first and only switch to more complex methods if the performance is considered insufficient - giving preference to more transparent, interpretable algorithms if possible. Finally, we present the best L2-regularization hyperparamters that we obtained for each of the 44 datasets, and encourage practitioners to use them for meta-learning in the low-data regime.

\subsubsection*{Acknowledgments}
This research was funded by the Bundesministerium für Bildung und Forschung (16DHBKI071, 01IS23041C).

\bibliography{iclr2024_conference}
\bibliographystyle{iclr2024_conference}

\newpage

\appendix
\section{Appendix} \label{sec:appendix}

\tabcolsep=0.13cm
\begin{longtable}{p{2cm} p{0.5cm} p{0.5cm} p{1.6cm} p{1.6cm} p{1.6cm} p{1.6cm} p{1.6cm} p{0.6cm}}
\caption{Training AUC (mean test AUC) and $\lambda^*$ across 44 PMLB datasets with sample size M and feature set size N, ordered for sample size ranges in steps of 100.}\\
\label{table:results}
\multirow{2}{=}[1em]{\bf PMLB dataset} & \multirow[t]{2}{=}{\bf M} & \multirow[t]{2}{=}{\bf N} & \multicolumn{2}{c}{\bf AutoML} & \multicolumn{2}{c}{\bf Deep learning} & \multirow{2}{=}{\bf Logistic regression} & \multirow[t]{2}{=}{$\boldsymbol{\lambda^*}$} \\
& & & \textbf{Auto-Prognosis} & \textbf{Auto-Gluon} & \textbf{TabPFN} & \textbf{HyperFast} & & 
\\ \hline \\
parity5 & 32 & 5 & 0.50 (0.27) & 0.04 (\textbf{0.98}) & 1.00 (0.02) & 1.00 (0.02) & 0.50 (0.17) & 0.5\\
analcatdata\_ fraud & 42 & 11 & 0.93 (\textbf{0.86}) & 0.99 (0.68) & 1.00 (0.79) & 0.99 (0.73) & 0.89 (0.77) & 0.5\\
analcatdata\_ aids & 50 & 4 & 1.00 (\textbf{0.73}) & 0.94 (0.67) & 1.00 (0.63) & 0.80 (0.53) & 0.78 (0.61) & 0.004\\
analcatdata\_ bankruptcy & 50 & 6 & 1.00 (\textbf{0.98}) & 1.00 (0.97) & 1.00 (0.96) & 0.99 (0.88) & 0.99 (0.97) & 0.004\\
analcatdata\_ japansolvent & 52 & 9 & 1.00 (0.85) & 0.99 (0.88) & 1.00 (\textbf{0.91}) & 0.97 (\textbf{0.91}) & 0.94 (0.85) & 0.1\\
labor & 57 & 16 & 1.00 (0.88) & 1.00 (0.95) & 1.00 (\textbf{0.99}) & 1.00 (0.98) & 1.00 (0.97) & 0.02\\
analcatdata\_ asbestos & 83 & 3 & 0.87 (\textbf{0.87}) & 0.89 (0.85) & 0.93 (0.85) & 0.87 (\textbf{0.87}) & 0.87 (0.86) & 0.5\\
lupus & 87 & 3 & 0.92 (0.84) & 0.86 (0.77) & 0.86 (0.82) & 0.83 (0.79) & 0.85 (\textbf{0.85}) & 0.1 \\
postoperative\_ patient\_data & 88 & 8 & 0.59 (\textbf{0.49}) & 0.12 (0.46) & 0.99 (0.44) & 0.87 (0.34) & 0.65 (0.38) & 0.5\\
analcatdata\_ cyyoung9302 & 92 & 10 & 1.00 (\textbf{0.89}) & 0.99 (0.85) & 0.99 (0.87) & 0.96 (0.84) & 0.94 (0.87) & 0.1\\
analcatdata\_ cyyoung8092 & 97 & 10 & 0.91 (0.73) & 0.99 (\textbf{0.87}) & 0.98 (0.85) & 0.91 (0.84) & 0.93 (0.79) & 0.1\\
analcatdata\_ creditscore & 100 & 6 & 1.00 (\textbf{1.00}) & 1.00 (0.99) & 1.00 (\textbf{1.00}) & 0.94 (0.87) & 0.97 (0.94) & 0.02\\
\\ \hline \\
\multicolumn{3}{l}{median M = 32, ..., 100} & 0.97 (\textbf{0.86}) & 0.99 (\textbf{0.86}) & 1.00 (0.85) & 0.95 (0.84) & 0.91 (0.85) & \\
\\ \hline \\
appendicitis & 106 & 7 & 0.88 (0.78) & 0.91 (0.85) & 0.97 (0.82) & 0.86 (\textbf{0.87}) & 0.86 (0.84) & 0.5\\
molecular\_bio-logy\_promoters & 106 & 57 & 1.00 (0.88) & 1.00 (\textbf{0.91}) & 1.00 (0.88) & 1.00 (0.89) & 1.00 (0.88) & 0.5\\
analcatdata\_ boxing1 & 120 & 3 & 0.97 (\textbf{0.89}) & 0.96 (0.85) & 0.99 (0.76) & 0.72 (0.67) & 0.68 (0.67) & 0.5\\
mux6 & 128 & 6 & 0.50 (\textbf{1.00}) & 1.00 (\textbf{1.00}) & 1.00 (\textbf{1.00}) & 1.00 (0.95) & 0.78 (0.70) & 0.5\\
analcatdata\_ boxing2 & 132 & 3 & 0.92 (\textbf{0.82}) & 0.91 (0.75) & 0.85 (0.71) & 0.75 (0.70) & 0.70 (0.68) & 0.5\\
hepatitis & 155 & 19 & 0.87 (\textbf{0.85}) & 0.99 (0.80) & 0.99 (\textbf{0.85}) & 0.92 (0.83) & 0.93 (0.84) & 0.5\\
corral & 160 & 6 & 1.00 (\textbf{1.00}) & 1.00 (\textbf{1.00}) & 1.00 (\textbf{1.00}) & 1.00 (\textbf{1.00}) & 0.97 (0.96) & 0.1\\
glass2 & 163 & 9 & 1.00 (0.89) & 1.00 (\textbf{0.91}) & 1.00 (0.89) & 0.89 (0.79) & 0.81 (0.72) & 0.004\\
backache & 180 & 32 & 0.90 (0.60) & 1.00 (0.71) & 1.00 (0.75) & 0.93 (\textbf{0.78}) & 0.90 (0.72) & 0.5\\
prnn\_crabs & 200 & 7 & 1.00 (\textbf{1.00}) & 1.00 (\textbf{1.00}) & 1.00 (\textbf{1.00}) & 0.83 (0.81) & 1.00 (\textbf{1.00}) & 0.004\\
\\ \hline \\
\multicolumn{3}{l}{median M = 101, ..., 200} & 0.95 (\textbf{0.89}) & 1.00 (0.88) & 1.00 (0.87) & 0.91 (0.82) & 0.88 (0.78)\\
\\ \hline \\
sonar & 208 & 60 & 1.00 (0.88) & 1.00 (\textbf{0.92}) & 1.00 (\textbf{0.92}) & 0.95 (0.89) & 0.95 (0.85) & 0.5\\
biomed & 209 & 8 & 1.00 (\textbf{1.00}) & 1.00 (0.96) & 1.00 (0.95) & 0.96 (0.93) & 0.96 (0.94) & 0.004\\
prnn\_synth & 250 & 2 & 0.98 (0.94) & 0.96 (\textbf{0.95}) & 0.97 (\textbf{0.95}) & 0.93 (0.94) & 0.94 (0.94) & 0.02\\
analcatdata\_ lawsuit & 264 & 4 & 1.00 (0.99) & 1.00 (0.99) & 1.00 (\textbf{1.00}) & 0.99 (0.98) & 1.00 (\textbf{1.00}) & 0.004\\
spect & 267 & 22 & 0.87 (\textbf{0.84}) & 0.94 (0.81) & 0.95 (0.83) & 0.90 (0.83) & 0.90 (0.82) & 0.5\\
heart\_statlog & 270 & 13 & 0.94 (\textbf{0.91}) & 0.97 (0.87) & 0.98 (0.90) & 0.93 (0.89) & 0.93 (0.89) & 0.5\\
breast\_cancer & 286 & 9 & 0.76 (0.69) & 0.96 (0.67) & 0.90 (\textbf{0.73}) & 0.80 (0.69) & 0.73 (0.70) & 0.5\\
heart\_h & 294 & 13 & 0.92 (0.87) & 0.96 (0.87) & 0.97 (\textbf{0.88}) & 0.89 (0.85) & 0.88 (0.86) & 0.5\\
hungarian & 294 & 13 & 0.99 (\textbf{0.86}) & 1.00 (0.85) & 0.96 (\textbf{0.86}) & 0.92 (0.84) & 0.89 (0.85) & 0.5\\
\\ \hline \\
\multicolumn{3}{l}{median M = 201, ..., 300} & 0.98 (0.88) & 0.97 (0.87) & 0.97 (\textbf{0.90}) & 0.93 (0.89) & 0.93 (0.86)\\
\\ \hline \\
cleve & 303 & 13 & 0.96 (\textbf{0.90}) & 0.98 (\textbf{0.90}) & 0.99 (0.89) & 0.93 (0.88) & 0.90 (0.88) & 0.5\\
heart\_c & 303 & 13 & 0.94 (\textbf{0.91}) & 0.95 (0.90) & 0.98 (\textbf{0.91}) & 0.94 (0.89) & 0.92 (\textbf{0.91}) & 0.5\\
haberman & 306 & 3 & 0.86 (0.70) & 0.76 (0.71) & 0.82 (\textbf{0.72}) & 0.65 (0.58) & 0.70 (0.66) & 0.5\\
bupa & 345 & 5 & 0.70 (0.66) & 0.77 (0.65) & 0.72 (\textbf{0.68}) & 0.72 (0.66) & 0.68 (0.67) & 0.1\\
spectf & 349 & 44 & 1.00 (0.91) & 1.00 (\textbf{0.94}) & 1.00 (0.93) & 0.91 (0.87) & 0.94 (0.88) & 0.1\\
ionosphere & 351 & 34 & 1.00 (0.97) & 1.00 (\textbf{0.98}) & 1.00 (\textbf{0.98}) & 0.96 (0.97) & 0.97 (0.90) & 0.5\\
colic & 368 & 22 & 0.99 (\textbf{0.87}) & 0.98 (\textbf{0.87}) & 1.00 (\textbf{0.87}) & 0.90 (0.86) & 0.89 (0.86) & 0.5\\
horse\_colic & 368 & 22 & 0.99 (\textbf{0.88}) & 0.98 (0.85) & 1.00 (0.84) & 0.89 (0.83) & 0.87 (0.82) & 0.5\\
\\ \hline \\
\multicolumn{3}{l}{median M = 301, ..., 400} & 0.98 (\textbf{0.89}) & 0.98 (\textbf{0.89}) & 1.00 (0.88) & 0.91 (0.87) & 0.90 (0.87)\\
\\ \hline \\
house\_votes\_84 & 435 & 16 & 1.00 (\textbf{0.99}) & 1.00 (\textbf{0.99}) & 1.00 (\textbf{0.99}) & 0.99 (0.98) & 0.99 (\textbf{0.99}) & 0.1\\
vote & 435 & 16 & 1.00 (\textbf{1.00}) & 1.00 (0.99) & 1.00 (\textbf{1.00}) & 0.99 (0.99) & 1.00 (0.99) & 0.5\\
saheart & 462 & 9 & 0.82 (\textbf{0.77}) & 0.69 (0.75) & 0.83 (\textbf{0.77}) & 0.81 (0.76) & 0.79 (\textbf{0.77}) & 0.5\\
clean1 & 476 & 168 & 1.00 (0.93) & 1.00 (\textbf{1.00}) & 1.00 (0.99) & 0.98 (0.96) & 1.00 (\textbf{1.00}) & 0.004\\
irish & 500 & 5 & 1.00 (\textbf{1.00}) & 1.00 (\textbf{1.00}) & 1.00 (\textbf{1.00}) & 0.98 (0.97) & 0.85 (0.83) & 0.1\\
\\ \hline \\
\multicolumn{3}{l}{median M = 401, ..., 500} & 1.00 (\textbf{0.99}) & 1.00 (\textbf{0.99}) & 1.00 (\textbf{0.99}) & 0.98 (0.97) & 0.99 (\textbf{0.99})\\
\\ \hline \\
\end{longtable}

\end{document}